\documentclass{article}

\usepackage{microtype}
\usepackage{graphicx}
\usepackage{subfigure}
\usepackage{booktabs} % for professional tables

\usepackage{hyperref}

\usepackage{url}
\usepackage{graphicx}

\usepackage{tikz}
\usepackage{blindtext}
\usepackage{hyperref}
\usepackage{url}
\usepackage{float}
\usepackage{multirow}
\usepackage{booktabs}
\usepackage{caption}
\usepackage{comment}
\usepackage{algorithmic}
\renewcommand{\algorithmiccomment}[1]{\hfill{$\triangleright$ #1}}
\usepackage{array}
\usepackage{wrapfig}
\usepackage{fdsymbol}
\usepackage{bbm}

\usepackage{nccmath} % this is used for reducing the spacing between text and equation.

\usepackage{wrapfig}

\usepackage{microtype}
\usepackage{graphicx}
\usepackage{booktabs} % for professional tables

\usepackage{hhline}
\usepackage{textcomp}

\definecolor{Note_color}{rgb}{0.0, 0.0, 1.0}

\usepackage[accepted]{icml2021}

\icmltitlerunning{Auto-NBA: Efficient and Effective Search Over the Joint Space of Networks, Bitwidths, and Accelerators}

\begin{document}

\twocolumn[

\icmltitle{Auto-NBA: Efficient and Effective Search Over the Joint Space of \\ Networks, Bitwidths, and Accelerators}



\icmlsetsymbol{equal}{*}

\begin{icmlauthorlist}
\icmlauthor{Yonggan Fu}{rice}
\icmlauthor{Yongan Zhang}{rice}
\icmlauthor{Yang Zhang}{ibm}
\icmlauthor{David Cox}{ibm}
\icmlauthor{Yingyan Lin}{rice}
\end{icmlauthorlist}

\icmlaffiliation{rice}{Department of Electrical and Computer Engineering, Rice University}
\icmlaffiliation{ibm}{MIT-IBM Watson AI Lab}

\icmlcorrespondingauthor{Yingyan Lin}{yingyan.lin@rice.edu}

\icmlkeywords{Machine Learning, ICML}

\vskip 0.3in
]

\printAffiliationsAndNotice{}  % leave blank if no need to mention equal contribution

\begin{abstract}

While maximizing deep neural networks' (DNNs') acceleration efficiency requires a joint search/design of three different yet highly coupled aspects, including the networks, bitwidths, and accelerators, the challenges associated with such a joint search have not yet been fully understood and addressed. The key challenges include (1) the dilemma of whether to explode the memory consumption due to the huge joint space or achieve sub-optimal designs, (2) the discrete nature of the accelerator design space that is coupled yet different from that of the networks and bitwidths, and (3) the chicken and egg problem associated with network-accelerator co-search, i.e., co-search requires operation-wise hardware cost, which is lacking during search as the optimal accelerator depending on the whole network is still unknown during search. To tackle these daunting challenges towards optimal and fast development of DNN accelerators, we propose a framework dubbed Auto-NBA to enable jointly searching for the \textbf{N}etworks, \textbf{B}itwidths, and \textbf{A}ccelerators, by efficiently localizing the optimal design within the huge joint design space for each target dataset and acceleration specification. Our Auto-NBA integrates a heterogeneous sampling strategy to achieve unbiased search with constant memory consumption, and a novel joint-search pipeline equipped with a generic differentiable accelerator search engine. Extensive experiments and ablation studies validate that both Auto-NBA generated networks and accelerators consistently outperform state-of-the-art designs (including co-search/exploration techniques, hardware-aware NAS methods, and DNN accelerators), in terms of search time, task accuracy, and accelerator efficiency. Our codes are available at: \href{https://github.com/RICE-EIC/Auto-NBA}{https://github.com/RICE-EIC/Auto-NBA}.

\end{abstract}
\vspace{-2em}
\section{Introduction}
\label{sec:intro}
The prohibitive complexity of deep neural networks (DNNs) has fueled a tremendous demand for efficient DNN accelerators which could boost the acceleration efficiency by orders of magnitude. 
In response, extensive research efforts have been devoted to developing DNN accelerators. 
Early works decouple the design of efficient DNN algorithms \cite{10.1145/3210240.3210337,pmlr-v80-wu18h,you2020shiftaddnet} and their accelerators ~\cite{du2015shidiannao,chen2017eyeriss,9138916,9138981}. On the algorithms level, pruning, quantization, or neural architecture search (NAS) have been adopted; On the hardware level, various FPGA-/ASIC-based accelerators customize the \textit{micro-architectures} (e.g., memory hierarchies/size and network-on-chip design) and algorithm-to-hardware \textit{mapping methods} (e.g., loop tiling strategies and loop orders) to optimize the acceleration efficiency for given DNNs. Later, hardware-aware NAS (HA-NAS) was proposed to further improve DNNs' acceleration efficiency \cite{tan2019mnasnet,fu2020autogandistiller}.   

It has been recently recognized that (1) optimal DNN accelerators require a joint consideration for three different yet coupled aspects: the network structure, network precision, and their accelerators, and (2) merely exploring a subset of these aspects will lead to sub-optimal hardware efficiency or task accuracy. For example, the optimal accelerators for DNNs with different structures (e.g., width/depth/kernel-size) can be very different; while the optimal networks and their bitwidths for different accelerators can differ a lot \cite{wu2019fbnet}.
However, the direction of jointly designing or searching for all three aspects has only been slightly touched on. For example,~\cite{chen2018joint, gong2019mixed, wang2020apq} proposed to jointly search for DNNs' structure and precision for a fixed target hardware;~\cite{abdelfattah2020best, yang2020co, jiang2020device, jiang2020hardware} proposed to jointly search for the networks and their accelerators, yet either their network or accelerator choices are limited, due to the prohibitive time cost required by their adopted reinforcement learning (RL) based methods; and EDD~\cite{li2020edd} formulated a differentiable joint search framework, which however only consider one accelerator parameter (i.e., parallel factor) and more importantly, has not yet solved the key challenges of efficient joint search. 

Although differentiable search is promising in terms of search efficiency to explore the huge joint search space (see Sec.~\ref{sec:dnas_motivation}), a plethora of challenges exist to achieve an effective, generic joint search for the above three aspects. First (\underline{\textit{Challenge} 1}), to co-search for a DNN and its precision, there exists a dilemma about whether to activate all the paths during search. On one hand, the required memory can easily explode and thus constrain the search scalability to more complex tasks if all paths are activated; on the other hand, partially activating a subset of the paths requires a sequential training of different precisions on the same weights, causing inaccurate accuracy ranking among different precisions~\cite{jin2020adabits}. 
Second (\underline{\textit{Challenge 2}}), DNN accelerators' design factors are not differentiable, and it is non-trivial to abstract generic accelerator design spaces integrating all important factors, e.g., the number of memory hierarchies, loop orders/sizes, and processing array size/shape.
Third (\underline{\textit{Challenge 3}}), there exists the chicken and egg problem associated with network-accelerator co-search, i.e., co-search requires operation-wise hardware costs, which is lacking during search as the optimal accelerator depending on the whole network is still unknown during search.

We aim to enable an efficient and effective joint search for the three aspects, and make contributions as follows:

\vspace{-0.5em}
\begin{itemize}
    \item We propose Auto-NBA that \textbf{for the first time} enables \textbf{Auto}mated joint search for the \textbf{N}etworks,  \textbf{B}itwidths, and \textbf{A}ccelerators
    for efficiently exploring the huge joint design space which cannot be afforded by previous RL-based methods due to their required prohibitive search cost. Auto-NBA identifies and tackles the above \textit{Challenges} 1-3 towards scalable joint search of the three for maximizing both the accuracy and efficiency. 

    \vspace{-0.3em}
    \item We propose a heterogeneous sampling strategy integrated by Auto-NBA for simultaneous updating the weights and network structures to (1) avoid  sequentially training different precisions and (2) achieve unbiased search with constant memory consumptions, i.e., solving the above \textit{Challenge} 1. We further develop a novel joint-search pipeline integrating a differentiable accelerator search engine to address \textit{Challenges} 2-3.
    
    \vspace{-0.3em}
    \item Extensive experiments and ablation studies validate the effectiveness and advantages of our Auto-NBA framework in terms of the resulting search time, task accuracy, and accelerator efficiency, when benchmarked over state-of-the-art (SOTA) co-search/exploration techniques, HA-NAS methods, and DNN accelerators, respectively. Furthermore, we visualize the searched accelerators by Auto-NBA to discuss insights towards efficient DNN accelerator design. 
    
     \vspace{-0.3em}
     \item Auto-NBA's searched algorithms and accelerators outperform both SOTA automatically searched and expert-designed DNNs and accelerators. Additionally, our Auto-NBA is general and allows users to easily plug-in both their own DNN search space and/or accelerator search space. Hence, we believe that Auto-NBA has made a nontrivial step to provide automated tools for expediting the development of DNN accelerators which falls far behind DNN algorithm advances.

\end{itemize}

\vspace{-1.3em}
\section{Related works}
\label{sec:related_work}

\textbf{Hardware-aware NAS.} 
Hardware-aware NAS (HW-NAS) automates the design of efficient DNNs. Early works~\cite{tan2019mnasnet, howard2019searching, tan2019efficientnet} utilize RL-based NAS that requires a massive search time/cost, while recent works~\cite{wu2019fbnet, wan2020fbnetv2, cai2018proxylessnas, stamoulis2019single} adopt differentiable search~\cite{liu2018darts} with much improved searching efficiency. Along another direction, one-shot NAS methods~\cite{cai2019once, guo2020single, yu2020bignas} pretrain the supernet and directly evaluate the performances of the sub-networks in a weight-sharing manner as a proxy of their independently trained performances at the cost of a longer pretrain time. Additionally, NAS has been adopted to search for quantization strategies~\cite{wang2019haq, wu2018mixed, cai2020rethinking, elthakeb2020releq} to trim down the complexity of DNNs. However, these works leave the hardware design space unexplored, which is a crucial enabler for DNN's acceleration efficiency.

\textbf{DNN accelerators.} Motivated by customized accelerators' large potential gains, SOTA accelerators~\cite{du2015shidiannao,chen2017eyeriss} innovate micro-architectures and mapping methods to optimize the acceleration efficiency, given a DNN and the hardware specifications. However, it is non-trivial to design an optimal accelerator as it requires cross-disciplinary \sloppy knowledge in algorithm, micro-architecture, and circuit design. SOTA accelerator design relies on either experts' time-consuming manual design or design flow~\cite{hls_chen2005xpilot,hls_chen2009lopass,hls_rupnow2011high} and DNN accelerator design automation~\cite{wang2016deepburning, zhang2018caffeine, guan2017fp, venkatesanmagnet,wang2018design, gao2017tetris,Xu_2020}. As they merely explore the accelerator space, they can result in sub-optimal solutions as compared to SOTA co-search/exploration methods and our Auto-NBA. 

\begin{figure*}[!t]
    \centering
    \includegraphics[width=0.85\textwidth]{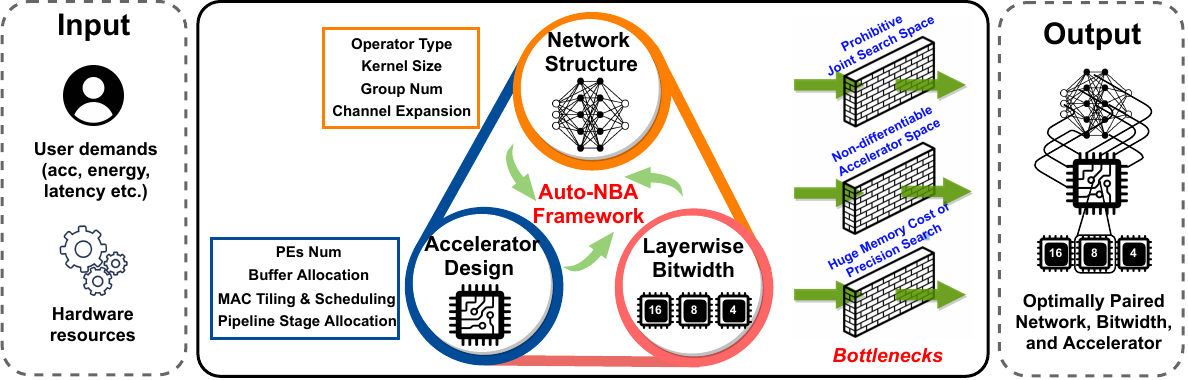}
 \vspace{-1em}
    \caption{Illustrating our Auto-NBA framework:  The middle part shows (1) a high-level view of Auto-NBA and (2) the technical challenges that Auto-NBA tackles for enabling a scalable, generic joint-search for the networks, bitwidths, and accelerators. }
    \label{fig:overview}
    \vspace{-1em}
\end{figure*}

\textbf{Co-exploration/search techniques.}
Pioneering efforts have been made towards jointly searching DNNs and their accelerators to some extent. 
For joint searching for DNNs and their precision,~\cite{chen2018joint, gong2019mixed, wang2020apq} adopt either differentiable or evolutionary algorithms yet without exploring their hardware accelerators.
For joint searching for DNNs and their accelerators,~\cite{abdelfattah2020best, yang2020co, jiang2020device, jiang2020hardware, jiang2020standing} conduct RL-based search for the networks and some accelerator parameters/templates, where they strictly constrain the search space of the network or accelerator to achieve a practical RL search time, limiting their scalability and achievable efficiency. \cite{linneural} attempts to co-design the network and accelerator in a sequential manner based on the fact that the accelerator’s design cycle is longer than the networks.
EDD~\cite{li2020edd} extends differentiable NAS to search for layer-wise precision and the accelerators' parallel factor, which is most relevant to our Auto-NBA. However, it has not yet solved the joint search challenges. First, it does not discuss or address the potentially explosive memory consumption issue of such a joint search; second, EDD's accelerator search space only includes one design parameter (i.e., the parallel factor), which is strictly limited to their accelerator template and cannot generalize to include common accelerator parameters such as the memory hierarchies and tiling strategies. 

Auto-NBA targets a scalable, generic joint-search framework 
for boosting the search efficiency and effectiveness.

\vspace{-0.5em}
\section{The Proposed Auto-NBA Framework}
\label{sec:methods}

In this section, we describe our proposed techniques for enabling Auto-NBA. Sec.~\ref{sec:trips_formulation} introduces Auto-NBA's formulation, while Sec.~\ref{sec:precision_search} and Sec.~\ref{sec:hardware_search} present Auto-NBA's technical enablers that address the key challenges of scalable joint search for the networks, bitwidths, and accelerators, and Sec.~\ref{sec:trips_cosearch} unifies the enablers to realize Auto-NBA.

\vspace{-0.5em}
\subsection{Auto-NBA: Problem Formulation}
\label{sec:trips_formulation}
Fig.~\ref{fig:overview} shows an overview of Auto-NBA, which jointly searches for the networks (e.g., kernel-size/channel-expansion/group-number), precision (e.g., 4-/6-/8-/12-/16-bit), and the accelerators (e.g., memory size and tiling strategies of each memory) in a differentiable manner. Auto-NBA targets a \textit{scalable} yet \textit{generic} joint search framework, which we formulate as a bi-level optimization problem: 

 \vspace{-1em}
\begin{align} %\label{eq:opt-df}
    \begin{split}
    & \quad \quad \quad \,\,\,\,\, \min \limits_{\alpha, \beta}  \,\, L_{val}( \omega^*, net(\alpha), prec(\beta)) \label{eq:update_alpha_beta} 
    \end{split} \\
    \begin{split}
    & s.t. \quad  L_{cost}(hw(\gamma^*), net(\alpha), prec(\beta)) \leq E_{target},  \label{eq:hw_cost}
    \end{split} \\
    \begin{split}
    & s.t. \quad \omega^* = \underset{\omega}{\arg\min} \,\, L_{train}( \omega,  net(\alpha), prec(\beta)),  \label{eq:update_weight}
    \end{split} \\
    \begin{split}
    & s.t. \quad \gamma^* = \underset{\gamma}{\arg\min} \,\, L_{cost}(hw(\gamma), net(\alpha), prec(\beta)) \label{eq:update_hw}
    \end{split}
\end{align}
\vspace{-2em}

where $\alpha$, $\beta$, and $\gamma$ are continuous variables parameterizing the probability of different choices for the network operators, precision bitwidths, and accelerator parameters as in~\cite{liu2018darts}, respectively; $\omega$ is the supernet weights; $L_{train}$, $L_{val}$, and $L_{cost}$ are the loss during training and validation, and the hardware-cost loss, respectively; $E_{target}$ is the target hardware cost (e.g., energy or latency); and $net(\alpha)$, $prec(\beta)$, and $hw(\gamma)$ denote the network, precision, and accelerator characterized by $\alpha$, $\beta$, and $\gamma$, respectively.

\begin{figure*}[!t]
    \centering
    \includegraphics[width=\textwidth]{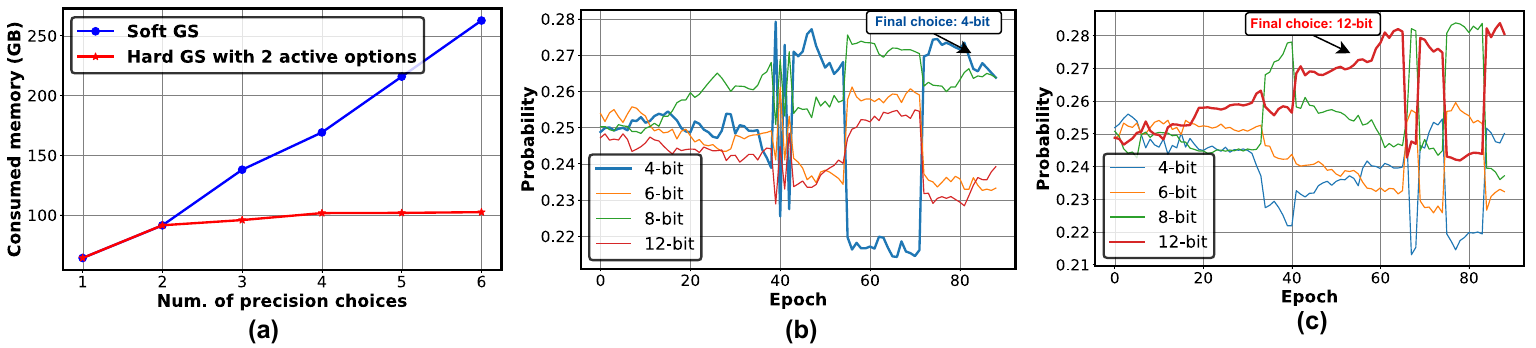}
    \vspace{-2.5em}
    \caption{(a) GPU memory consumption comparison between soft Gumbel Softmax (GS) and hard GS sampling, which are two activating approaches for co-search for the network and precision; and the probability evolution of each precision choice during the search process in the 4-th block when searching with: (b) hard GS sampling for updating both the weights $\omega$ and precision choices $\beta$, which results in the lowest 4-bit, and (c) the proposed heterogeneous sampling for updating $\omega$ and $\beta$, which results in the highest 12-bit (desired).}
    \label{fig:precision_search}
    \vspace{-1em}
\end{figure*}

\vspace{-0.5em}
\subsection{Auto-NBA Enabler 1:  Heterogeneous Sampling 
 for Scalable Network-Precision Joint-Search}
\label{sec:precision_search}

As discussed in Sec.~\ref{sec:intro}, there exists a dilemma (i.e., either memory explosion or biased search) whether to activate all the paths during precision search, for tackling which we propose a simple yet effective heterogeneous sampling strategy. Here we first use real experiments to illustrate the joint-search dilemma and then introduce our heterogeneous sampling which effectively addresses the challenge.

\textbf{Activating all choices $\rightarrow$ memory explosion and entangled correlation among choices.} During precision search, activating all the precision choices as~\cite{wu2018mixed, gong2019mixed} can easily explode the memory consumption especially when the precision is co-searched with the network structures. While composite convolutions~\cite{cai2020rethinking} for mixed-precision search can potentially mitigate this memory explosion issue during search by shrinking the required computation, yet the large memory consumption issue would still exist during training when updating the precision parameters, i.e., $\beta$ in Eq.~(\ref{eq:update_alpha_beta}). For example, as shown in Fig.~\ref{fig:precision_search} (a), the measured GPU memory consumption of co-searching for the network and precision on ImageNet grows linearly with the number of precision choices if activating all precision choices during search. Furthermore, the entangled correlation (e.g., co-adaptation~\cite{hong2020dropnas}, correlation~\cite{li2019stacnas}, and cooperation~\cite{tian2020discretization}) among different precision choices can lead to a large gap between the supernet during search and the final derived network, thus failing the joint search.

\textbf{Activating only a subset of choices - Biased search.} For addressing the above issues of memory explosion and correlation among choices, one natural approach is to adopt hard Gumbel Softmax (i.e., activating only one or a subset of paths as~\cite{dong2019searching}) to constrain the memory consumption, which however can lead to a biased search and thus poor performance. Specifically, activating only a subset of the precision choices implies a sequential training of different precisions that can lead to inaccurate performance ranking. This is because a sequential training means different precision choices are applied on top of the same weights and activations. As a result, different precision choices can interfere with each other, and different training orders would lead to different results. For a better understanding, we next show two concrete experiments.  

\begin{table}[!t]
 \vspace{-0.5em}
  \centering
  \caption{Comparing the accuracy when training a fixed network using different precision schedules, where high2low and low2high denote progressive training from high precision to low precision and the inverse case, respectively, following~\cite{jin2020adabits}.}
%   \vspace{-0.5em}
    \resizebox{0.5\textwidth}{!}{
\begin{tabular}{c||ccccc}
% \toprule
\toprule
\multirow{2}{*}{\textbf{Strategy}} &  \multicolumn{5}{c}{\textbf{Resulting Accuracy}} \\ 
  &  4-bit (\%) & 8-bit (\%) & 12-bit (\%) & 16-bit (\%) &  32-bit (\%) \\ \midrule
\textbf{Independent Train} & 63.52 & 67.44 & 67.56 & 67.65 & 68.21 \\ \midrule
\textbf{high2low Train} & 59.29 & 45.09 & 45.45 & 45.15 & 65 \\ 
\textbf{low2high Train} & 4.36 & 26.55 & 43.58 & 63.3 & 63.5 \\ \midrule
\textbf{joint Train} & 63.28 & 66.98 & 67.21 & 67.23 & 67.36 \\
\bottomrule 
\end{tabular}
    }
  \label{tab:training_strategy}%
  \vspace{-1.5em}
\end{table}%

\textit{Co-search network and precision using hard Gumbel Softmax:} Fig.~\ref{fig:precision_search} (b) shows the resulting precision probability evolution when co-searching for the network and precision on CIFAR-100 using hard Gumbel Softmax, which activates two precision choices, without imposing any hardware-cost constraints, thus the desired and effective precision choice would be the highest precision. However, as shown in Fig.~\ref{fig:precision_search} (b), the block co-searched using hard Gumbel Softmax collapses to the lowest precision (i.e., the highest probability towards the end of the search is the lowest precision choice 4-bit), indicating an ineffective search direction. Note that the fluctuation in the probability of different precision choices is caused by the intermittent activation of the block due to the hard Gumbel Softmax sampling.

\textit{Sequential training of a fixed network with multiple precision choices:}
As observed in~\cite{jin2020adabits}, when training a fixed network with multiple precision choices, either ascending or descending the precision will incur an inferior convergence and thus chaotic accuracy ranking among different precision choices. For example, as shown in Tab.~\ref{tab:training_strategy}, we compare the accuracy of a fixed network (all blocks adopt the k3e1 (kernel size 3 and channel expansion 1) structure in~\cite{wu2019fbnet}) under different precision choices, when being trained with different precision schedule strategies. We can see that only jointly training all the precision choices can maintain a ranking consistent with that of independently trained ones, while sequential training leads to both inferior accuracy and ranking. 

\textbf{Proposed solution - Heterogeneous sampling.} To tackle both aspects of the aforementioned dilemma, we propose 
a heterogeneous sampling strategy as formulated below: 

\useshortskip
\vspace{-0.5em}
\begin{align}
    \begin{split}
    \hspace{-0.8em}A^{l+1} &= \bar{W}^l * \sigma(\bar{A}^l) = \sum_{j=1}^J \bar{\beta^l_j} W_j^l * \sigma(\sum_{j=1}^J \bar{\beta^l_j} A_j^l) \,\,\,\, \\
    \hspace{-0.8em}where \,\,
     \bar{\beta^l_j} &=
    \begin{cases}
      GS(\beta^l_j) & \text{if updating weights}\\
      GS_{hard}(\beta^l_j) & \text{if updating $\beta$}
    \end{cases} 
    \end{split}  \label{eq:hetero_search}
\end{align}
\vspace{-1.5em}

where $\bar{W}^l$ / $\bar{A}^l$ are the composite weights/activations of the $l$-th layer as in~\cite{cai2020rethinking} which are the weighted sum of weights/activations under different precision choices, e.g., $W_j^l$ is the weights quantized to the $j$-th precision among the total $J$ options for the $l$-th layer, and $\sigma$ is the activation function.

Our heterogeneous sampling updates the weights in Eq.~(\ref{eq:update_weight}) by jointly updating the weights under all the precision choices weighted by their corresponding soft Gumbel Softmax $GS(\beta^l_j)$, where $\beta^l_j$ parameterizes the probability of the $j$-th precision in the $l$-th layer, and the gradients is estimated by the straight-through estimator (STE)~\cite{zhou2016dorefa} as $\partial L_{train}/\partial A^l \approx \partial L_{train}/\partial \bar{A^l}$ so that no extra intermediate feature maps need to be stored during backward. For updating $\beta$, we adopt hard Gumbel Softmax~\cite{jang2016categorical} with one-hot outputs $GS_{hard}(\beta^l_j)$ to save memory computation while reducing the correlation among precision choices. Under the same co-search setting as Fig.~\ref{fig:precision_search} (b), all the blocks searched using our heterogeneous sampling converge to the highest precision towards the end of the search (see Fig.~\ref{fig:precision_search} (c)), indicating an effective search as further validated  in Sec. \ref{sec:exp}.

\vspace{-0.5em}
\subsection{Auto-NBA Enabler 2: Differentiable Accelerator Search Engine}
\label{sec:hardware_search}
\textbf{Motivation.} Although EDD~\cite{li2020edd} also co-searches the accelerator with the network, their search space is limited to include merely one accelerator parameter (i.e., the parallel factor within their template) which can be fused into their computational cost, whereas this is not always applicable to other naturally non-differentiable accelerator design parameters such as memory hierarchies and tiling strategies. Hence, a more general and efficient search engine is needed towards generic differentiable accelerator search.

\textbf{Search algorithm.} We propose a differentiable search engine to efficiently search for the optimal accelerator (including the micro-architectures and mapping methods) given a DNN and its precision using single-path sampling as discussed in Sec.~\ref{sec:trips_cosearch}. Specifically, we solve Eq.~(\ref{eq:update_hw}) as follows: 

\useshortskip
\vspace{-0.5em}
\begin{equation} \label{eqn:hw_diff}
    \begin{split}
    % \gamma^* = \,\, 
    \underset{\gamma}{\arg\min} \,\, \sum_{m=1}^{M} GS_{hard}(\gamma^m) \, L_{cost}(hw(\{GS_{hard}(\gamma^m)\}), \\ net(\{O_{fw}^l\}), prec(\{B_{fw}^l\}))\\
    \end{split}
\end{equation}
\vspace{-1.7em}

\noindent where $M$ is the total number of accelerator design parameters. Given the network $net(\{O_{fw}^l\})$ and precision $prec(\{B_{fw}^l\})$, 
% \NOTE{where $O_{fw}^l \in \alpha$ and $B_{fw}^l \in \beta$ are the only operator and precision activated during forward as discussed in Sec.~\ref{sec:trips_cosearch}},
where $O_{fw}^l \in \alpha$ and $B_{fw}^l \in \beta$ are the activated forward operator and precision for each layer as discussed in Sec.~\ref{sec:trips_cosearch}.
Our search engine utilizes hard Gumbel Softmax $GS_{hard}$ sampling on each design parameter $\gamma^m$ to build an accelerator $hw(\{GS_{hard}(\gamma^m)\})$ and penalize each sampled accelerator  parameter with the overall hardware-cost $L_{cost}$ through relaxation in a gradient manner.

\textbf{Hardware template.} We adopt a unified template for both the FPGA and ASIC accelerators, which is a parameterized chunk-based pipeline micro-architecture inspired by~\cite{shen2017}. As elaborated in Sec. \ref{sec:setup}, the hardware/micro-architecture template comprises multiple sub-accelerators (i.e., chunks) and executes DNNs in a pipeline fashion. Each chunk is assigned multiple but not necessarily consecutive layers which are executed sequentially within the chunk. Similar to Eyeriss, each chunk consists of several levels of memories (e.g., on-chip buffer and local register files) and processing elements (PEs) to facilitate data reuses and parallelism with searchable design knobs, such as PE interconnections (i.e., Network-on-chip), allocated buffer sizes, multiply-and-accumulate (MAC) operations’ scheduling and tiling (i.e., dataflows), and so on.

\textbf{General applicability.} As shown in Eq.~(\ref{eqn:hw_diff}), our accelerator search engine is general and does not hold any prior assumptions about the adopted accelerators. Hence, it is applicable to different accelerator architectures and mapping methods. Specifically, for a given target accelerator architecture or template, such as TPU-like~\cite{TPU} or other accelerators \cite{chen2016eyeriss,9138916,9138981}, our search engine can be directly applied once given (1) a simulator to estimate the hardware cost, and (2) a set of user-defined searchable accelerator design knobs abstracted from the target accelerator template.

\vspace{-0.5em}
\subsection{Auto-NBA: The Overall Joint-Search Framework}
\label{sec:trips_cosearch}

\textbf{Objective and challenges.} 
\underline{The key objective} of Auto-NBA is formulated in Eq.~(\ref{eq:update_alpha_beta}) involving all the three major aspects towards efficient DNN accelerators.
\underline{The key challenges} for joint-search of the three include \textbf{(1)} the prohibitively large joint space (e.g., \textbf{2.3E+21} in this work) which, if not addressed, will limit the search scalability to practical yet complex tasks; \textbf{(2)} the entangled co-adaptation~\cite{hong2020dropnas}, correlation~\cite{li2019stacnas}, and cooperation~\cite{tian2020discretization} issues among different network and precision choices can enlarge the gap between the supernet during search and the final derived network, thus failing the joint search; and \textbf{(3)} the chicken and egg problem associated with network-accelerator co-search, i.e., co-search requires operation-wise hardware cost, which is lacking during search as the optimal accelerator depending on the whole network is still unknown during search.

\begin{table*}[!t]
  \centering
  % \vspace{-1em}
  \caption{
Benchmark Auto-NBA's Search efficiency over SOTA co-search/exploration works and one-shot NAS methods.}
%   \vspace{-0.3em}
    \resizebox{\linewidth}{!}{
\begin{tabular}{ccccccc}
\toprule
Method          & Dataset   & Network Space & Accelerator Space & Precision Space & Joint Space       & Search Time (GPU hours) \\ 
\midrule \midrule
HS-Co-Opt~\cite{jiang2020hardware}       & CIFAR-10  & 1.15E+18      & -                 & -               & 1.15E+18          & 103.9                   \\
\textbf{Auto-NBA}           & CIFAR-10  & 9.85E+20      & 2.24E+27          & 2.40E+15        & \textbf{5.30E+63} & \textbf{6}              \\
\midrule
BSW~\cite{abdelfattah2020best}    & CIFAR-100 & 4.20E+05      & 8.64E+03          & -               & 3.63E+09          & 5184                    \\
\textbf{Auto-NBA}           & CIFAR-100 & 9.85E+20      & 2.24E+27          & 2.40E+15        & \textbf{5.30E+63}          & \textbf{12}                      \\
\midrule
HS-Co-Opt~\cite{jiang2020hardware}       & ImageNet  & 2.22E+18      & -                 & -               & 2.22E+18          & 266.8                   \\
Once-For-All~\cite{cai2019once}    & ImageNet  & 2.00E+19      & -                 & -               & 2.00E+19          & 1200                    \\
APQ~\cite{wang2020apq}             & ImageNet  & 1.00E+35      & -                 & 1.00E+10        & 1.00E+45          & 2400                    \\
Single One-shot~\cite{guo2020single} & ImageNet  & 7.00E+21      & -                 & -               & 7.00E+21          & 288                     \\
\textbf{Auto-NBA}           & ImageNet  & 9.85E+20      & 2.24E+27          & 2.40E+15        & \textbf{5.30E+63}          & \textbf{80}        \\
\bottomrule
\end{tabular}
    }
  \label{tab:search_efficiency}
  \vspace{-1em}
\end{table*}

\textbf{Auto-NBA implementation.} Auto-NBA integrates the two enablers in Sec.~\ref{sec:precision_search} and Sec.~\ref{sec:hardware_search} to develop a unified joint-search pipeline. Specifically, Auto-NBA search starts from updating both the supernet weights $\omega$ and accelerator parameters $\gamma$ (based on Enablers 1-2 in Sec.~\ref{sec:precision_search} and Sec.~\ref{sec:hardware_search}, respectively), given the current network $net(\alpha)$ quantized using precision $prec(\beta)$, and then updates $\alpha$ and $\beta$ based on the derived optimal weights $\omega^*$ and accelerator $hw(\gamma^*)$ resulting from the previous step. 

During joint-search, Auto-NBA updates $\alpha$ and $\beta$ as follows (see Eq.~(\ref{eq:forward})-Eq.~(\ref{eq:backward_cost})) to solve Eq.~(\ref{eq:update_alpha_beta}), where only the update for $\alpha$ is shown for simplicity as it is similarly applicable to update $\beta$. Note that here we define \textit{path} to be one of the parallelled candidate operators between the layer input and layer output within one searchable layer, which can be viewed as a coarse-grained (layer-wise) version of the path definition in~\cite{wang2018interpret,qiu2019adversarial}.

\textit{Single-path forward:} For updating both $\alpha$ (see Eq.~(\ref{eq:forward})) and $\beta$ during forward, Auto-NBA adopts hard Gumbel Softmax sampling~\cite{hu2020dsnas}, i.e., only the choice with the highest probability will be activated to narrow the gap between the search and evaluation, leveraging the single-path property of hard Gumbel Softmax sampling. In Eq.~(\ref{eq:forward}), $A^{l}$ and $A^{l+1}$ denote the feature maps of the $l$-th and $(l+1)$-th layer, respectively, $N$ is the total number of operator choices, $O_i^l$ is the $i$-th operator in the $l$-th layer parameterized by $\alpha_i^l$, and $O_{fw}^l$ is the activated operator during forward.

\useshortskip
\vspace{-0.5em}
\begin{align}
\small
    \begin{split}
    &Forward: A^{l+1} = \sum_{i=1}^{N} GS_{hard}(\alpha_i^l) \, O_i(A^l)
    = O_{fw}^l(A^l)
    \label{eq:forward} 
    \end{split} \\
\small
    \begin{split}
    &Backward: \frac{\partial L_{val}}{\partial \alpha_i^l} = \sum^{K}_{k=1} \frac{\partial L_{val}}{\partial GS(\alpha_k^l)} \frac{\partial GS(\alpha_k^l)}{\partial \alpha_i^l} \\
   & \quad \quad \quad \quad \quad \quad \quad \quad = \frac{\partial L_{val}}{\partial A^{l+1}} \sum^{K}_{k=1} O_k^l(A^l) \frac{\partial GS(\alpha_k^l)}{\partial \alpha_i^l}
    \label{eq:backward_val} 
    \end{split} \\
\small
    \begin{split}
           & \frac{\partial L_{cost}}{\partial \alpha_i^l} = \mathbbm{1}(GS_{hard}(\alpha_i^l)=1) \, L^{\alpha_i^l}_{cost}(hw(\gamma^*), net(\alpha_i^l), prec(\beta))
    \label{eq:backward_cost} 
    \end{split}
\end{align}
\vspace{-1.5em}

\textit{Multi-path backward:} For updating both $\alpha$ (see Eq.~(\ref{eq:backward_val})) and $\beta$ during backward, Auto-NBA activates multiple paths to calculate the gradients of $\alpha$ and $\beta$ through Gumbel Softmax relaxation in order to balance the search efficiency and stability inspired by~\cite{cai2018proxylessnas, hu2020tf}, where $\alpha_i^l$'s gradients are calculated using Eq.~(\ref{eq:backward_val}), with $K\in(1,N)$ being the number of activated choices with the top $K$ Gumbel Softmax probability and controlling the search cost.

\textit{Hardware-cost penalty:} The network search in Eq.~(\ref{eq:update_alpha_beta}) is performed in a layer/block-wise manner as in~\cite{liu2018darts}, thus requiring layer/block-wise hardware-cost penalty which is determined by both the layer/block-to-accelerator \textit{mapping method} and the corresponding layer/block execution cost on the optimal accelerator $hw(\gamma^*)$. The optimal mapping method is yet determined by the whole network. To handle this gap, we derive the layer/block-wise hardware-cost 
assuming that the single-path network derived from the current forward would be the final derived network, as this single-path network has a higher if not the highest probability to be finally derived. In Eq.~(\ref{eq:backward_cost}), $\mathbbm{1}(\cdot)$ is an indicator denoting whether $\alpha_i^l$ (i.e., the $i$-th operator in the $l$-th layer) is activated during forward. 
\vspace{-0.5em}
\section{Experiment Results}
\label{sec:exp}
\subsection{Experiment Setup}
\label{sec:setup}
\textbf{Software settings.} 
\underline{Search space and hyper-params.} 
We adopt the same search space as~\cite{wu2019fbnet} for the ImageNet experiments, from which we disable the first two down-sampling operations for the CIFAR-10/100 experiments. We use [4, 6, 8, 12, 16] as the candidate precision set, where the precisions of the first and last blocks are fixed to 8-bit, and each block shares the same precision for both the weights and activations for more hardware-friendly implementation. We activate two paths during backward, i.e., $K=2$ in Eq.~(\ref{eq:backward_val}), for search efficiency. For $L_{cost}$ in Eq. (~\ref{eq:update_hw}), we use the acceleration latency, i.e., Frame-Per-Second (FPS), and Energy-Delay-Product (EDP) for FPGA- and ASIC-based accelerators, respectively.

\underline{Search settings.} 
We adopt standard search settings used in SOTA hardware-aware NAS works \cite{wu2019fbnet}. Specifically, for searching on the CIFAR-10/100 datasets, we use half of the dataset for updating supernet weights $\omega$ and the other half for updating the network and precision parameter $\alpha$ and $\beta$, and search for 90 epochs with an initial gumbel softmax temperature of 5 decayed by a factor of 0.975 every epoch; For searching on ImageNet, we randomly sample 100 classes as a proxy search dataset from which we use 80\% for updating $\omega$ and the other 20\% for updating $\alpha$ and $\beta$, pretrain the supernet for 30 epochs without updating the network architecture and precision, and then search for 90 epochs with an initial temperature of 5 decayed by a factor of 0.956 every epoch, following~\cite{wu2019fbnet}. For both CIFAR-10/100 and ImageNet, we use an initial learning rate of 0.1 and an annealing cosine learning rate.

\underline{Training settings.} 
For CIFAR-10/100, we train the derived networks for 600 epochs using a batch size of 256 with an initial learning rate of 0.1 and an annealing cosine learning rate on a single NVIDIA RTX-2080Ti GPU following~\cite{liu2018darts}. For ImageNet, we follow the training recipe in~\cite{wu2019fbnet} on four NVIDIA Tesla V100 GPUs.

\underline{Baselines.}
We benchmark against four kinds of SOTA baselines: (1) the most relevant baseline EDD~\cite{li2020edd} which co-searches for networks, precisions, and one accelerator paremeters, (2) SOTA methods co-exploring networks and accelerators including HS-Co-Opt~\cite{jiang2020hardware}, NASAIC~\cite{yang2020co}, BSW~\cite{abdelfattah2020best}, and NHAS~\cite{linneural}, (3) SOTA methods co-searching for the network and precision including APQ~\cite{wang2020apq} and MP-NAS~\cite{gong2019mixed}, and (4) hardware-aware NAS with uniform precision, including FBNet~\cite{wu2019fbnet}, ProxylessNAS~\cite{cai2018proxylessnas}, and Single-Path NAS~\cite{stamoulis2019single}.

\vspace{-0.5em}
\textbf{Hardware settings.}
\underline{Search space.}
Our accelerator search space is inspired by a SOTA accelerator architecture ~\cite{shen2017, dnnbuilder} and adopts a chunk-wise pipelined architecture, aiming to more efficiently accelerate more recent networks which have diverse network structures. Specifically, our accelerator search space is a parameterized chunk-wise pipelined architecture~\cite{shen2017, dnnbuilder}, in which the following parameters are searchable: \textbf{(1)} the parallel PE array, i.e., the number and the inter-connections of the PEs, \textbf{(2)} the on-chip buffers, i.e., allocated lower-level memories for the inputs, weights, and outputs, \textbf{(3)} the tiling and scheduling for the MAC computations, and \textbf{(4)} the network layer allocation, i.e., how to assign each layer to be processed by different chunks within the chunk-wise pipelined architecture, with all being critical accelerator parameters as pointed out by SOTA accelerator works~\cite{chen2017eyeriss, zhang2015, yang2016systematic}. 
To facilitate automated search, all the choices for the aforementioned accelerator parameters are formatted and maintained using vectors so that they can be compatible with the optimization formulation in Sec.~\ref{sec:hardware_search}. Note that users of our proposed Auto-NBA can easily plug in their preferred accelerator search space as discussed in Sec.~\ref{sec:hardware_search}.

\underline{Platforms.} To evaluate the generated network and accelerator designs, for FPGA accelerators, we adopt the standard Vivado HLS~\cite{vivado_HLS} design flow on the target Xilinx ZC706 development board~\cite{zc706}, which has a total $900$ DSPs (Digital Signal Processors) and $19.1$Mb BRAM (Block RAM); for ASIC accelerators, we use the SOTA energy estimation tool Timeloop~\cite{timeloop} and Accelergy,~\cite{accelergy}, to validate our generated design's performance, with CACTI7~\cite{cacti7} and Aladdin~\cite{aladdin} at a 32nm CMOS technology as unit energy and timing cost plugins.

\vspace{-0.5em}
\subsection{Auto-NBA vs. SOTA in Search Efficiency}
\label{sec:dnas_motivation}

To evaluate the superiority of Auto-NBA in terms of search efficiency, we compare the search space size and search time of Auto-NBA with both RL-based co-search/exploration works and one-shot NAS methods using the reported data from the baselines’ original papers as shown in Tab.~\ref{tab:search_efficiency}. We can see that Auto-NBA consistently requires a notably less search time while handling the largest joint search space on all the considered tasks. In particular, compared with the one-shot NAS methods~\cite{guo2020single, cai2019once} which can be potentially extended to implement co-search yet can suffer from a large pretraining cost, Auto-NBA achieves a 3.6$\times$ $\sim$ 30$\times$ less search time on ImageNet, justifying our choice of differentiable co-search.

\begin{figure}[!t]
    \centering
    % \vspace{-0.5em}
    \includegraphics[width=0.4\textwidth]{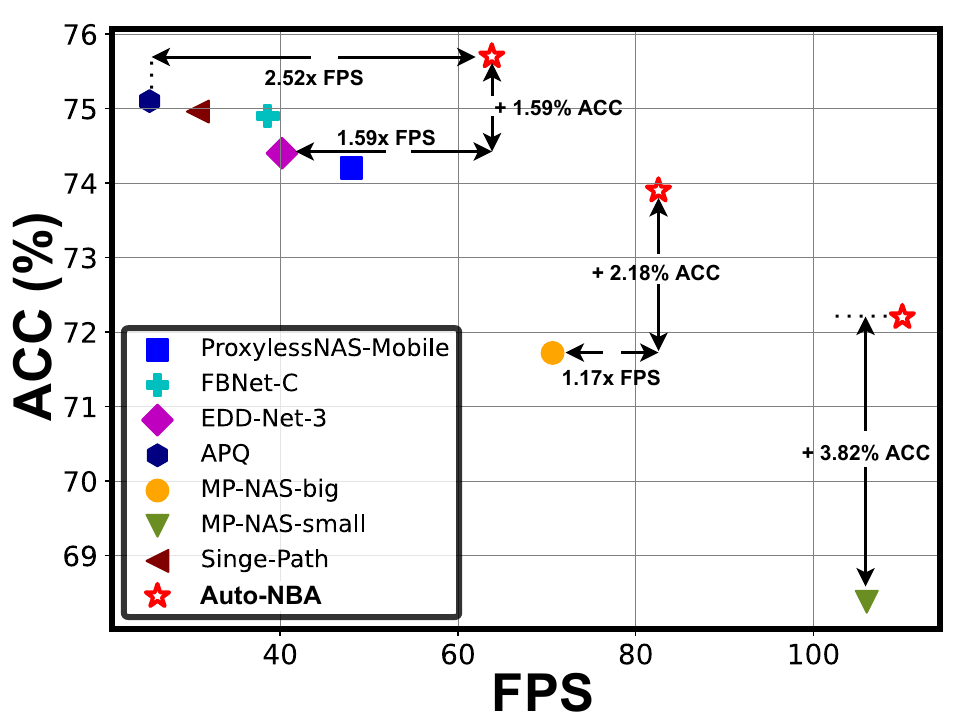}
    \vspace{-1.3em}
    \caption{Accuracy vs. FPS trade-off of Auto-NBA against SOTA efficient DNN solutions on ImageNet.}
    \label{fig:sota}
    \vspace{-2em}
\end{figure}

\begin{figure*}[htb]
    \centering
    % \vspace{-1em}
    \includegraphics[width=0.9\textwidth]{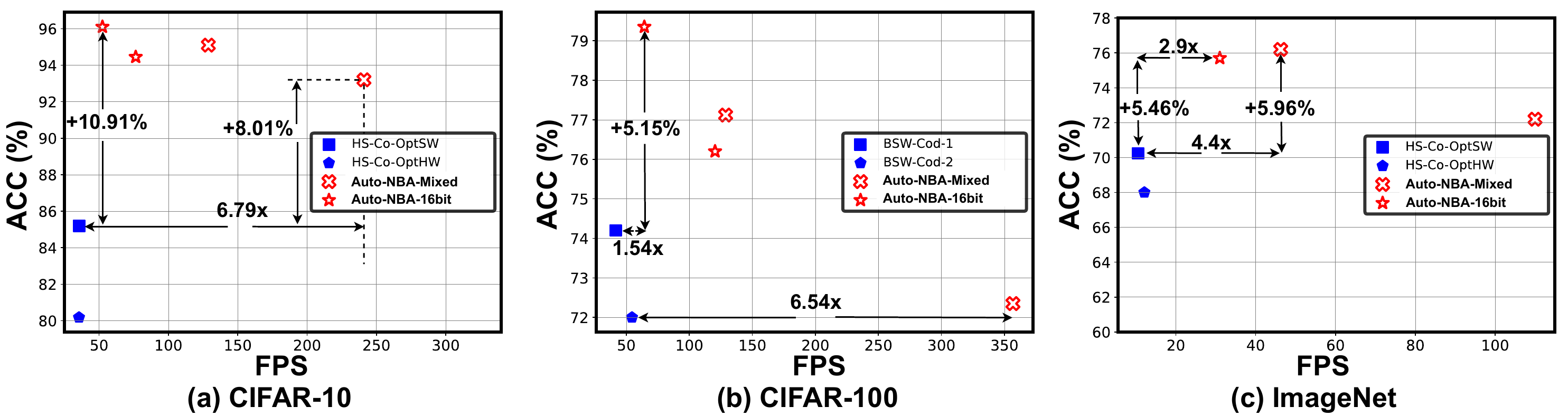}
    \vspace{-1em}
    \caption{Benchmark Auto-NBA w/ and w/o precision search (denoted as Auto-NBA-Mixed and Auto-NBA-16bit, respectively) with SOTA network/accelerator co-exploration methods~\cite{jiang2020hardware, abdelfattah2020best} on CIFAR-10/100/ImageNet.}
    \label{fig:benchmark_fpga}
     \vspace{-1em}
\end{figure*}

\vspace{-0.5em}
\subsection{Auto-NBA vs. SOTA in Searched Accelerators}
\textbf{Co-exploration of networks, precision, and accelerators.}
Here we benchmark Auto-NBA with SOTA automatically searched, expert designed, and co-searched/co-explored DNN algorithms/accelerators on ImageNet, considering FPGA-based accelerators as shown in Fig.~\ref{fig:sota} which include four Auto-NBA searched results for a fair comparison. 
We can observe that (1) the searched networks by our Auto-NBA consistently push forward the frontier of accuracy-FPS trade-offs, compared to all SOTA baselines, and (2) compared with the most relevant baseline EDD, we achieve a +1.3\% higher accuracy together with a 1.59$\times$ better FPS. The consistently large improvement of Auto-NBA over SOTA methods in co-design/co-exploration validates the necessity and effectiveness of Auto-NBA joint-search for all three aspects towards efficient DNN accelerators.

Note that we use EDD's reported results, and search for the optimal accelerator based on our accelerator space for APQ, MP-NAS, and SOTA hardware-aware NAS methods; the ProxylessNAS-8bit result is reported in APQ~\cite{wang2020apq}; and the other baselines are all quantized to 8-bit for hardware measurement and the accuracies are from the original papers without considering their accuracy degradation due to quantization effects. All methods consider a 450 DSP limit in FPGA for a fair comparison.

\textbf{Co-exploration of networks and accelerators.}
Software-Hardware co-design is a significant property of our Auto-NBA framework, so we further benchmark Auto-NBA with both searched precision and fixed-precision over SOTA network/accelerator co-search/exploration works.

\begin{table}[!t]
   \vspace{-1em}
  \centering
  \caption{Comparing the accuracy and ASIC efficiency (i.e., EDP and area) of Auto-NBA and SOTA co-exploration ASIC works~\cite{yang2020co}.
}
%   \vspace{-0.5em}
    \resizebox{0.5\textwidth}{!}{
\begin{tabular}{c||ccc}
\toprule
Optimization & Accuracy & EDP & Area \\
Methods & (\%) & (J*clock-cycle) & ($um^2$) \\\midrule
NAS $\rightarrow$ ASIC & 94.17 & 3.30E+06 & 4.83E+09 \\ 
ASIC $\rightarrow$ HW-NAS & 92.53 & 2.81E+06 & 3.86E+09 \\
NASAIC & 92.62 & 1.62E+06 & 3.34E+09 \\ \midrule
\textbf{Auto-NBA} & \textbf{94.34} & \textbf{4.36E+03} & \textbf{5.92E+05} \\ \bottomrule
\end{tabular}
    }
  \label{tab:asic}%
  \vspace{-1.5em}
\end{table}%

\underline{Co-search on FPGA.} We benchmark with HS-Co-Opt~\cite{jiang2020hardware} and BSW~\cite{abdelfattah2020best} on ZC706, under the same DSP limits as the baselines on CIFAR-10/100/ImageNet. Since all the baselines adopt a 16-bit fixed-point design, here we provide Auto-NBA with both fixed 16-bit and searched precision for a fair comparison. From Fig.~\ref{fig:benchmark_fpga}, we can see that (1) on both CIFAR-10/100, Auto-NBA with fixed 16-bit consistently achieves a better accuracy (up to 10.91\% and 5.15\%, respectively) and a higher FPS (up to 2.21$\times$ and 2.15$\times$, respectively) under the same DSP constraint, and (2) when co-searching for the precision, Auto-NBA can more aggressively push forward the FPS improvement (up to 6.79$\times$ and 6.54$\times$, respectively on CIFAR-10/100), implying the importance of co-exploring the precision dimension in addition to network and accelerator co-explorations. Specifically, Auto-NBA with searched precision achieves a +5.96\% higher accuracy and 4.4$\times$ FPS improvement on ImageNet over~\cite{jiang2020hardware}.
%reuse v1 with our results new 

\begin{table}[!b]
  \vspace{-2.5em}
  \centering
  \caption{Benchmark Auto-NBA over NHAS~\cite{linneural} under the same precision setting.
}
%   \vspace{-0.5em}
\resizebox{0.49\textwidth}{!}{
\begin{tabular}{cccc}
\hline
\textbf{Co-search} & \textbf{Accuracy} & \textbf{Latency} & \textbf{Area} \\
\textbf{Methods} & \textbf{(\boldmath{$\%$)}} & \textbf{\boldmath{($ms$)}} & \textbf{\boldmath{($mm^2$)}}  \\\hline \hline
NHAS~\cite{linneural} & 70.74 & 1.58 & 5.87  \\
\textbf{Auto-NBA} & \textbf{71.70} & \textbf{1.25} & \textbf{5.50} \\
\hline
\end{tabular}
}
\label{tab:asic_extra}
 \vspace{-1em}
\end{table}

\begin{figure}[!h]
    \centering
  \vspace{-0.5em}
    \includegraphics[width=0.35\textwidth]{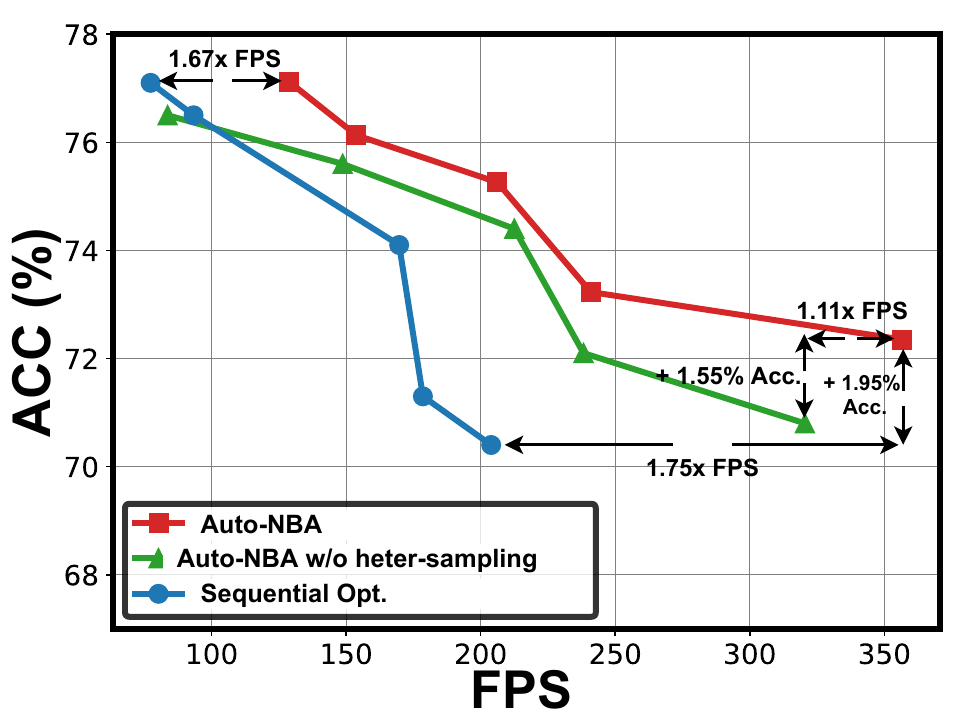}
    \vspace{-1.3em}
    \caption{Accuracy vs. FPS trade-off of Auto-NBA, Auto-NBA w/o heterogeneous sampling, and the sequential optimization baseline on CIFAR-100, under an FPGA DSP limit of 512.
    \label{fig:ablation_search}}
    \vspace{-2.5em}
\end{figure}

\begin{figure*}[!t]
    \centering
    % \vspace{-2em}
    \includegraphics[width=0.9\textwidth]{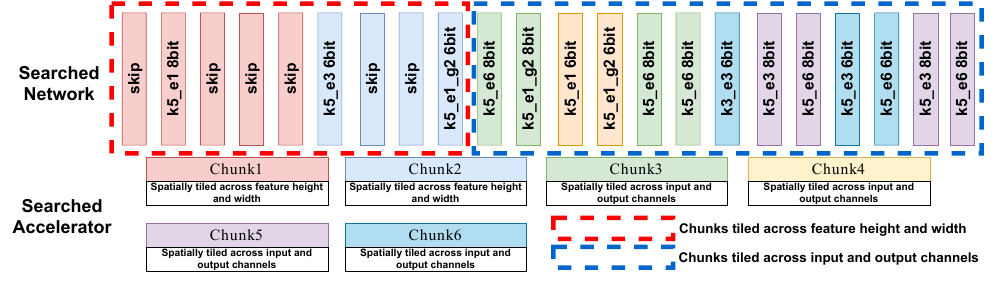}
    \vspace{-2em}
    \caption{Visualization of the searched network, precision, and accelerator that achieves a 72.2\% top -1 accuracy on ImageNet and an FPS of 110 on ZC706 FPGA, where the block definition follows~\cite{wu2019fbnet}.}
    \label{fig:visualization}
    \vspace{-1em}
\end{figure*}

\underline{Co-search on ASIC.} Here we evaluate Auto-NBA against three SOTA co-search methods for ASIC-based accelerators.
In Tab.~\ref{tab:asic}, we benchmark Auto-NBA with NASAIC~\cite{yang2020co} on CIFAR-10, which is the first exploration towards network/accelerator co-search targeting ASIC accelerators, considering both their reported co-search, sequential optimization, and hardware-aware optimization results. We can observe that compared with both co-search, sequential optimization, and hardware-aware optimization methods for exploring the ASIC-based accelerator design space, our Auto-NBA consistently achieves notably improved trade-offs between accuracy and EDP, which is equal to the acceleration energy cost multiplied with the acceleration latency (a commonly used metric for ASIC-based accelerators). In particular, Auto-NBA achieves a +0.17\% $\sim$ +1.81\% higher accuracy together with a 371.56$\times$  $\sim$ 756.88$\times$ reduction in EDP. In the baseline implementations~\cite{yang2020co}, most of the area is occupied by the support for heterogeneous functionalities, which leads to severely low utilization of the PE arrays when executing one task, thus leading to a surprisingly higher area and energy consumption.

We further benchmark Auto-NBA over another co-search baseline for ASIC-based accelerators, i.e., NHAS~\cite{linneural}. In particular, we fix the precision of Auto-NBA to be 4-bit for a fair comparison. As shown in Tab.~\ref{tab:asic_extra}, Auto-NBA achieves a 0.96\% higher accuracy and a 20.9\% reduction in latency under a comparable area consumption compared with NHAS, verifying the superiority of our Auto-NBA.

\vspace{-0.5em}
\subsection{Auto-NBA: Ablation Studies}
\textbf{Scalability under the same DSP.} Fig.~\ref{fig:ablation_search} shows the pareto frontier achieved by Auto-NBA under the same DSP constraint with different accuracy and FPS trade-offs on CIFAR-100, which indicates that Auto-NBA can handle and is scalable to a large range of required acceleration efficiency.

\textbf{Effectiveness of heterogeneous sampling.} In addition to the example and analysis in Sec.~\ref{sec:precision_search}, we further 
validate the effectiveness of the proposed \textit{heterogeneous} sampling strategy by benchmarking Auto-NBA w/ and w/o  \textit{homogeneous} sampling that adopts hard GS sampling ($K=2$) for updating both the weights $\omega$ and precision choices $\beta$ as that in Fig.~\ref{fig:precision_search} (b), the latter of which is termed as Auto-NBA w/o h-sampling. The achieved trade-offs between the task accuracy and acceleration FPS in Fig.~\ref{fig:ablation_search} show that
Auto-NBA w/o h-sampling tends to select lower precision choices which hurt the achieved accuracy, and is consistently inferior to Auto-NBA with heterogeneous sampling, due to its inaccurate estimation for different precision ranking.

\textbf{Comparison with sequential optimization.}
Considering the great flexibility on both DNNs' structure and accelerator sides, a natural baseline of Auto-NBA is to search the network and precision based on theoretical efficiency metrics (e.g., total bit operations), and then search for the optimal accelerator given the searched network and precision from the first search. We benchmark Auto-NBA over the aforementioned sequential search in Fig.~\ref{fig:ablation_search} on CIFAR-100, which shows that Auto-NBA consistently outperforms the sequential optimization baseline, e.g., a 1.95\% higher accuracy with a 1.75$\times$ better FPS, indicating the poor correlation between theoretical efficiency and real hardware efficiency and thus motivating the necessity of joint-search.

\subsection{Visualization of the searched network, precision, and accelerator}
\label{sec:visualization}

Fig.~\ref{fig:visualization} visualizes Auto-NBA' searched network, precision, and accelerator, from which we discuss our extracted insights below.

\textbf{Insights for the searched networks of Auto-NBA.} 
The automatically searched network of Auto-NBA is shown in Fig.~\ref{fig:visualization} and we can find that wide-shallow networks tend to better favor real-device efficiency on the ZC706 FPGA board while achieving a similar accuracy. We conjecture the reason is that wider networks offer more opportunities for making use of feature/channel-wise parallelism for a given batch size, thus leading to a higher resource utilization rate and thus an overall higher throughput.

\textbf{Insights for the searched accelerators of Auto-NBA.} 
As shown in Fig.~\ref{fig:visualization}, we can observe that the whole network is partitioned into multiple pipelined chunks to maximize the acceleration throughput, with each chunk being highlighted using a different color. As~\cite{shen2017} points out, such multi-chunk accelerator architectures can boost the overall utilization of the PE arrays via 1) optimizing each accelerator chunk (i.e., sub-accelerator) for a cluster of layers that have similar patterns/workloads and 2) pipelining all the chunks to process different network inputs and process non-consecutive layers.
Additionally, the chunks which are assigned with the early layers of the network prefer spatially tiling the feature map height and width as this offers more parallelism, while the chunks handling the deeper layers of the network tend to tile the channel dimension as the parallelism opportunity is more prominent along channel dimensions at the deeper layers. 

An ablation study for Auto-NBA's accelerator search engine is provided in the Appendix.

  \vspace{-0.8em}
\section{Conclusion}
  \vspace{-0.5em}
 In this work, we present Auto-NBA, which is the first to identify and tackle the prohibitive challenges of jointly search for the networks, bitwidths, and accelerators for maximizing task accuracy and acceleration efficiency. When benchmarking with a comprehensive set of SOTA efficient DNN algorithms, accelerators, and co-explored/co-searched works, Auto-NBA consistently achieves large improvements, outperforming both SOTA automatically searched and expert-designed DNNs and accelerators. Auto-NBA promises to expedite the development of DNN accelerators which falls far behind DNN algorithm advances.

\section*{Acknowledgements}
The work is supported by the National Science Foundation (NSF) CAREER Program (Award number: 2048183).

\nocite{langley00}

\bibliography{ref}
\bibliographystyle{icml2021}

\end{document}